# Hybrid Cryptocurrency Pump and Dump Detection

Hadi Mansourifar
hmansourifar@uh.edu
University of Houston
Houston, Texas

Lin Chen
Lin.Chen@ttu.edu
Texas Tech University
Lubbock, Texas

Weidong Shi
wshi3@central.uh.edu
University of Houston
Houston, Texas

## ABSTRACT

Increasingly growing Cryptocurrency markets have become a hive for scammers to run pump and dump schemes which is considered as an anomalous activity in exchange markets. Anomaly detection in time series is challenging since existing methods are not sufficient to detect the anomalies in all contexts. In this paper, we propose a novel hybrid pump and dump detection method based on distance and density metrics. First, we propose a novel automatic thresh-old setting method for distance-based anomaly detection. Second, we propose a novel metric called density score for density-based anomaly detection. Finally, we exploit the combination of density and distance metrics successfully as a hybrid approach. Our experiments show that, the proposed hybrid approach is reliable to detect the majority of alleged P & D activities in top ranked exchange pairs by outperforming both density-based and distance-based methods.

## 1 INTRODUCTION

Cryptocurrency pump and dump [1,14,15,16] is a fraudulent price manipulation in which coin traders will gather to plot a scam to encourage group of investors to buy a specific coin to push the price higher and a sudden dump to victimize gullible traders. Pump and dump has a long history in financial systems. However, social media allow the traders to plan short term market manipulation activities via targeting larger variety of victims [17]. To do so, traders declare a date, time, exchange market and a target coin to execute the pump phase. It 's not more than a quick scam to excite traders which takes place in minutes. According to a report published in the Wall Street Journal [2] a pump and dump scam took place in a group with more than 74,000 followers on the messaging app Telegram. The scam encouraged the followers to start purchasing a relatively unknown crypto coin called CloakCoin. As a consequence, the price of Cloak-Coin surged on the Binance exchange market, causing $ 1.7 million increase in trades. The price of CloakCoin raised 50 percent and two minutes later dropped by almost $ 1. The price of the other coins on the Binance did not change at the same time. The market for cryptocurrencies [11,12,18,19,20] is rapidly expanding. For instance, the Bitcoin [13] as top cryptocurrency had a market capitalization of more than 144 billion US dollars at the time of writing this paper [3]. The surging amounts of trades and investments in cryptocurrencies attracts groups of frenzy traders who want to gain profit in minutes. These investors are easy targets for scams designed for gullible traders. That 's why intelligent pump and dump detection and probability prediction systems are becoming known as neces-sary auxiliary tools to help financial systems to tackle the rising challenges in this area. Unsupervised anomaly detection is sup-posed a viable solution to detect suspicious activities in exchange markets. However, applying anomaly detection in time series has proven to be challenging. One reason is difficulty of setting multiple thresholds. Furthermore, traditional non-hybrid methods can not claim superiority in all contexts. As reported in [4], a traditional anomaly detection may easily fail to detect alleged pump and dump activities in exchange data. In this paper, we propose a novel hybrid anomaly detection method by combination of distance and density metrics. We show that, some outliers are detected by distance-based method and some by density-based method with little overlap. A hybrid approach however, can take advantage of both with promising anomaly detection success rates. To the best of our knowledge, such a hybrid approach based on distance and density metrics has not been investigated in the literature. We take the first step in this direction to combine distance and density metrics for pump and dump detection. Afterwards, we investigate the impact of each mentioned metric by analyzing Density-Distance distribution diagram.

our contributions are as follows.

- We propose a novel method to transfer contextual anomaly problem into point anomaly problem.
- We propose a novel automatic threshold setting method based on histogram processing for distance-based point anomaly detection.
- We propose a novel metric called density score for density-based anomaly detection.
- Our experiments show that, a hybrid approach based on dis-tance and density metrics can detect the majority of alleged pump and dump activities in top ranked exchange pairs.

The organization of the paper is as follows. Section 2, reviews related work in pump and dump detection and anomaly detection. Section 3, demonstrates the proposed method. Section 4 presents the experiments and finally section 5 concludes the paper.

## 2 RELATED WORK

Pump and dump detection researches can be divided into two differ-ent categories: Trade-based and Information-based. In this section, we briefly review the previous works related to each one. We also provide a short background about anomaly detection which is used in Trade-based P & D detection researches.

### 2.1 Trade-based Pump and Dump Detection

In trade-based P & D detection researches , sequence of trades in different exchange markets are analyzed to detect the anomalous sequences within available records. This type of researches do not take into account any auxiliary information to link external events and sudden changes in exchange markets. Leangarun and

Tangamchit et. Al [5] trained neural network models to detect trade-based stock price manipulations using the Level 2 data.

Level 1 data include history of buy/sell orders that are success-fully executed. level 2 data show an order ID, or buyer/seller ID, which can be an important clue to detect patterns of purchase be-longing to an individual. In general, level 2 data are private and it can only be accessible by market authorities. Kamps and Kleinberg [4] utilized contextual anomaly detection to detect the P & D schemes across various exchange markets. However, their proposed method failed to pass all test cases due to limitations of contextual anomaly detection. Here, we briefly review required backgrounds about anomaly detection.

2.1.1 *Anomaly Detection.* Anomaly detection [6,8] is the process of finding abnormal patterns or unexpected instances. Anomalies can be divided into three categories:
- point anomalies [25]: It 's the simplest anomaly category and a data point is considered anomalous if it 's located too far from the centroid of all data. A real-world application is credit card fraud detection based on finding anomalous spending patterns.
- collective anomalies [6]: Collective anomalies are a set of point anomalies which are linked to each other. Here, the goal is to detect community of point anomalies.
- contextual anomalies [9,10]: This type is also known as con-ditional anomalies and often occurred in time series data. To detect contextual anomalies, we need to monitor a period of continuous sequences. A contextual anomaly detection algorithm needs to set several thresholds related to various features to detect suspicious activities. Therefore, detecting contextual anomaly by judging time series plot is difficult since it needs to more than one contextual attribute.

It is hard to make a general algorithm for contextual anomaly since a normal incident in a context may be considered abnormal in another context. That's why various anomaly detection methods have been proposed for different applications. Contextual anomaly problems are normally handled in one of two ways: 1- Transforming a contextual anomaly detection into a point anomaly detection. 2-analyzing all sequences concurrently to detect anomalous records.

## 2.2 Information-Based Pump and Dump Detection

Information- based P & D detection methods investigate the role of online chat rooms to plan P & D schemes. Researches in this area are rare since the topic has been investigated very recently. First study of this type done by Xu and Livshits [7]. They took advantage of two different sources of data:
- Price and volume data belonging to multiple crypto exchanges.
- Telegram groups dedicated to pump and dump activities.

## 3 PROPOSED METHOD

Contextual anomaly detection is challenging because what is con-sidered anomalous today may be appeared normal in the future. Most of the events in financial systems can be interpreted in the same volatile atmosphere as they are influenced by variety of fac-tors. One viable solution to address contextual anomaly problem is to transform it to point anomaly problem. In this section, we propose a novel method to transform contextual anomaly problem into point anomaly problem for cryptocurrency P & D detection and probability prediction.

## 3.1 Pump and Dump Detection

To detect the P & D in time series data, first, we divide time series data into a set of different frames. Second, we concatenate all the data inside each frame to form a single high dimensional data per window. Given all virtually generated high dimensional data, the task is to pinpoint whether or not a data representing a frame is
outlier. To simplify the issue, Principal Component Analysis (PCA) is used to project the high dimensional data into 2D. Algorithm 1 shows the required steps.

**Algorithm 1** Contextual To Point Anomaly Detection (D,W) **Input**: Time series data D; Window size W
**Output**: Transformed data **D´**

1: Given **D**, **W**, divide **D** into $\frac{Size(\mathbf{D})}{\mathbf{W}}$ different frames and save them in **F**.
2: Given **F**, Concatenate data points inside each frame to extract one high-dimensional data point per frame and save them in **F´**.
3: Given **F'**, project extracted data points into 2D using Principle Component Analysis (PCA).
4: Save projected data in **D´**.
5: Return **D´**.

After projecting each frame into 2D using PCA, any point anom-aly detection algorithm can be used to detect the outliers as frames representing the P & D events.
We use following point anomaly detection approaches:
- Distance-based point anomaly detection: Projected data points are categorized into normal and anomalous based on distance to global centroid.
- Density-based point anomaly detection: Projected data points are categorized into normal and anomalous based on their density score.
- Hybrid Approach: Both density and distance-based methods are used to vote if a data point is anomalous.

## 3.2 Distance based anomaly detection

In this approach, the global centroid of projected data is calculated. Afterwards, anomalous data points are detected using a threshold as described in Algorithm 2. Point anomalies may appear in variable distances on different exchange pairs with respect to global centroid. Thus, we need an automatic threshold setting method.

## 3.3 Automatic threshold setting for point anomaly detection

Threshold setting in time series is challenging since it needs to set multiple thresholds for several features. Transforming the contextual anomaly detection into point anomaly detection using pro-posed method reduces the number of required thresholds to only

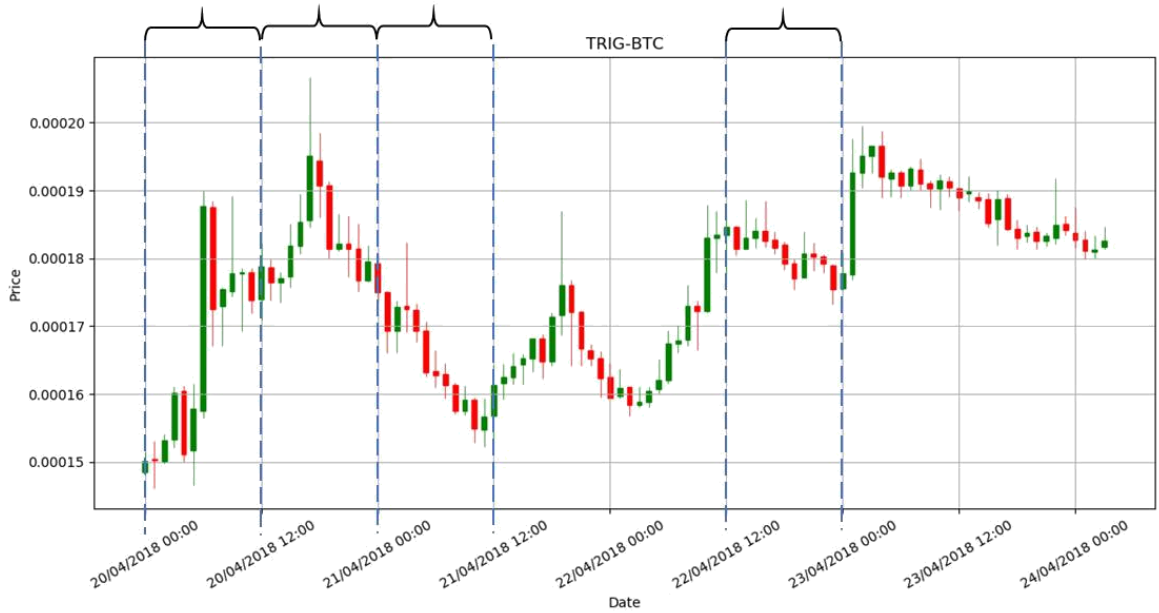

Figure 1: Dividing the time series data into frames to extract one high dimensional data per each window.

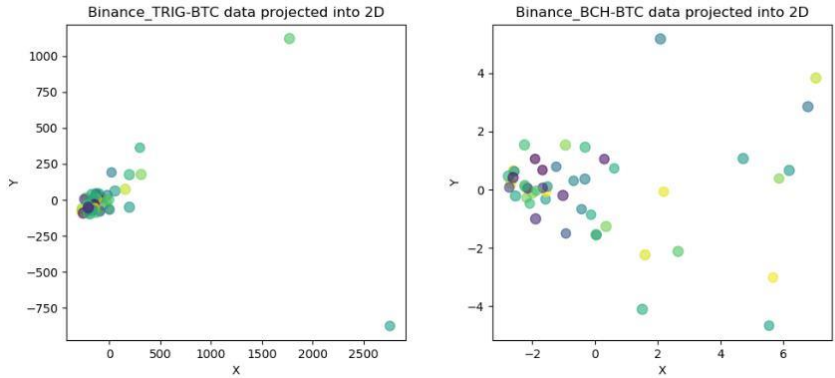

Figure 2: Projecting the time series data into 2D using proposed method. Outliers in TRIG-BTC are more conspicuous than BCH-BTC.

**Algorithm 2** Distance-based point Anomaly Detection (**D´,T**) Input: Projected data **D´**; Threshold **T** Output: Outlier data **O**

1: Given **D´**, Calculate the centroid of all instances as **C.**
2: Given **C**, calculate the distance of each instance of **D´** to **C** and save in dis.
3: Given **D´**, **dis**, **C**, **T** find the instances where $dis_i > T$ and save them in **O**.
4: Return **O**.

one. However, setting a single threshold is still difficult. To address this problem, we process the distance histogram of data calculated with respect to global centroid. Our experiments show that, the distance histogram can be categorized into two sections: dense region and sparse region. Our observations show that, sparse re-gion starts immediately after first empty histogram element. For example, suppose following sequence as histogram data.

[4, 10, 21, 4, 3, 0, 2, 0, 0, 0, 1, 0, 0, 0, 0, 0, 0, 0, 0, 0, 1, 0, 0, 1, 0, 0, 0, 1] The first zero element appears in sixth element. So we can divide the histogram as follows.

Dense region: [4, 10, 21, 4, 3]

Sparse region: [2, 0, 0, 0, 1, 0, 0, 0, 0, 0, 0, 0, 0, 0, 1, 0, 0, 1, 0, 0, 0, 1] Sparse and dense regions can be interpreted as safe and alleged P & D zones as shown in figure 4.

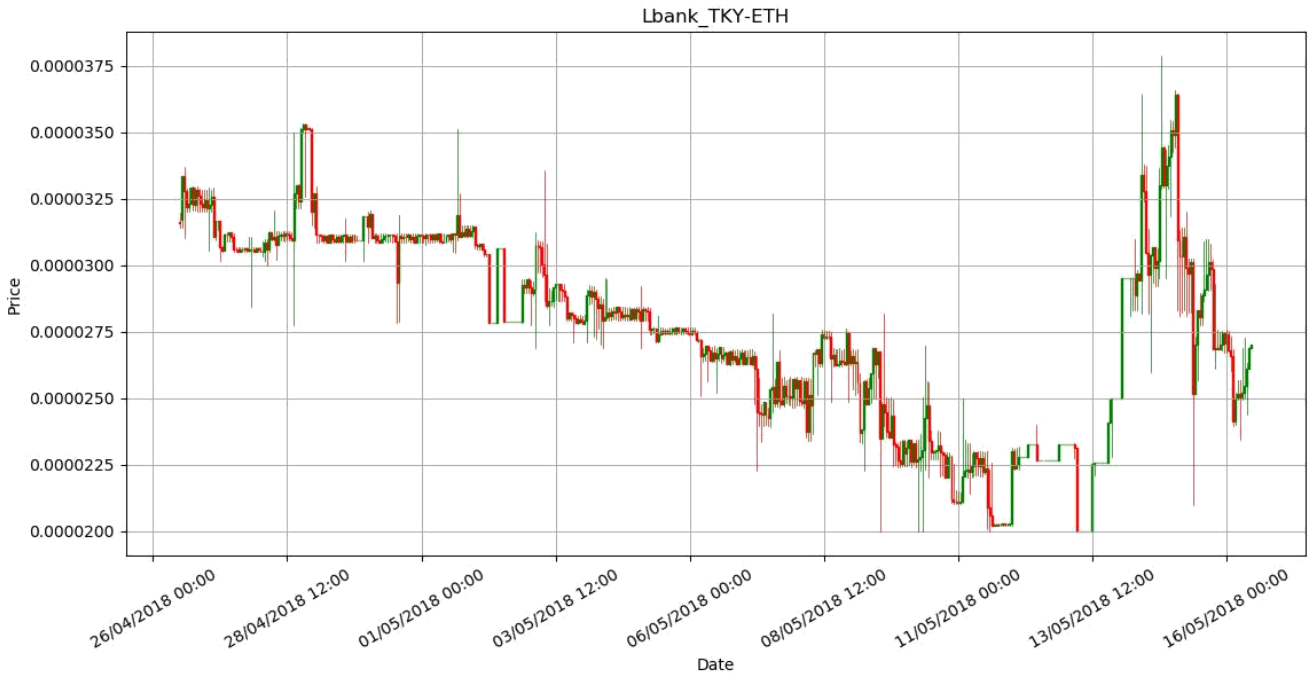

**Figure 3: Candlestick chart for the TKY/ETH trading pair. It 's difficult to detect the threshold level from this chart.**

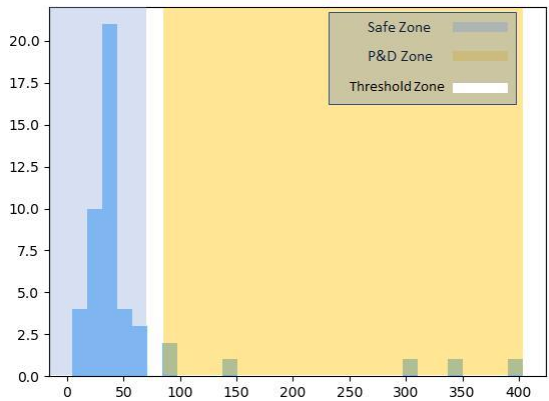

**Figure 4: Automatic threshold setting by histogram process-ing. The histogram is divided to sparse and dense regions.**

## 3.4 Density based point anomaly detection

In this section, we propose a novel metric called density score for density-based anomaly detection. Instead of finding the density of a region, density score is calculated for each data point. It helps us to find anomalous points located near the centroid which can not be detected by distance-based point anomaly detection. To find the density score of each instance, n nearest neighbors of each data point are found and saved in a common list. Afterwards, the density score of each data point is calculated based on its frequency in mentioned common list. Algorithm 3 shows the required steps to find density score of each data point.

---

**Algorithm 3** Density based Point Anomaly Detection (**D',n**) Input: Projected data **D´**; Number of nearest neighbor n Output: Anomalous data points **A**

---

1: Given **D´**, Calculate the n nearest neighbor of all instances and save them in **U**.
2: Given **U**, calculate the frequency of each instance in U as den-sity score and save them in **Fr**.
3: Given **Fr**, if *density score < 2* label it as anomalous and save in **A.**
4: Return **A.**

---

## 3.5 Hybrid point anomaly detection

Hybrid approaches have been proven to be efficient for anomaly detection [23,24]. Although distance-based anomaly detection and density-based anomaly detection are robust to detect anomalous points, none of them can detect all outliers. Our observations show that, some outliers can be detected by density-based method and some by distance-based method. To take advantage of both, a hybrid method can be used as a voting approach to find anomalous points. To the best of our knowledge such hybrid approach has not reported in the literature. Figure 5 shows Density-Distance distribution of projected data: TRIG/BTC diagram is stretched in X direction. It means that it's more sensitive to distance. However, BCH/BTC is stretched in Y direction meaning that it's sensitive to density.

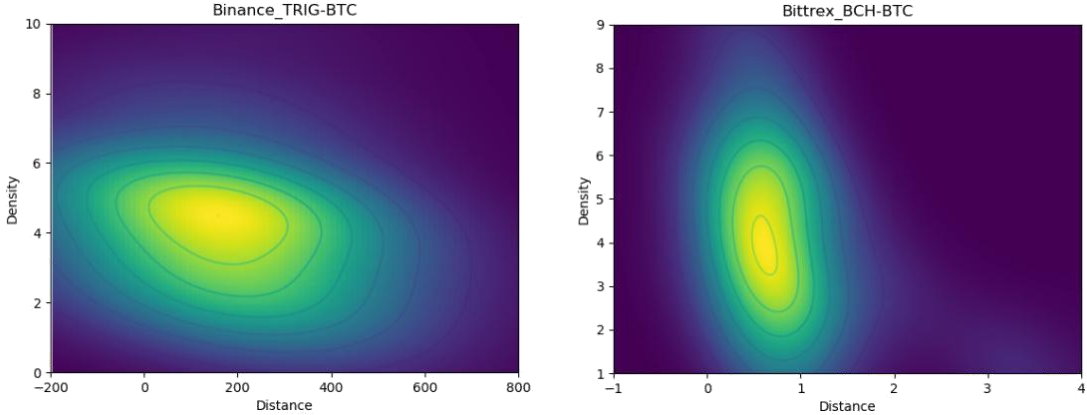

**Figure 5: Density-Distance distribution diagram. TRIG-BTC is stretched in X direction (distance sensitive) and BCH-BTC is stretched in Y direction (density sensitive).**

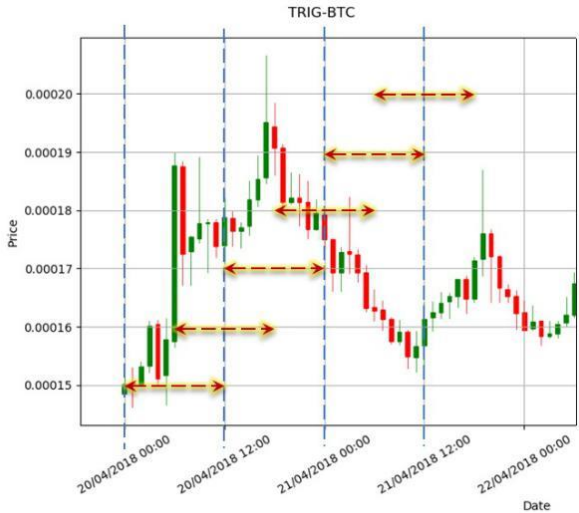

**Figure 6: Shifting frames are used for P & D prediction.**

## 3.6 Pump and Dump Prediction

Transforming contextual anomaly problem to point anomaly prob-lem using proposed method has a significant advantage: we can use the same technique but with different logic for P & D prediction. The main differences between P & D detection and prediction are as follows.

- In P & D detection a data point is extracted per window and we look for outliers among them. However, in P & D prediction, we keep track of only one data point which is shifted gradually.
- In P & D detection, there is no any overlap between adjacent frames but in P & D prediction there is a predefined overlap between adjacent frames as shown in figure 6.

To predict a P & D event we extract a data point from the first frame by the same approach as mentioned in Algorithm 1. Afterwards we shift the frame 1 such that frame 1 ∩ frame2 ≠ Null. Algorithm 4 shows the required steps.

---

**Algorithm 4** Cryptocurrency Pump and Dump Prediction (**D,W,T,Sh**)

**Input**: Time series data **D**; Window size **W**; Threshold **T**; Shift size **Sh.**
**Output**: Alleged frame $F_o$

1: Given **D, W**, select the first frame and concatenate data points inside it to extract one high-dimensional data point, project it into 2D and save it in **F**.
2: Given **Sh**, Shift the frame, concatenate the data points inside it to extract one high-dimensional data point, project it into 2D and save it in **Fn**.
3: Given **F, Fn** as initial frame and new frame, calculate the dis-tance between them and store it in dis .
4: If *dis>T* Return $F_o$.
5: Go to Step 2 until all the data are processed.

---

Figure 7 shows the distance diagram used for P & D prediction.

## 4 EXPERIMENTS

In this section, we introduce the dataset used for the sake of experi-ments. We also test the proposed method on top 10 symbols ranked by number of alleged pump and dump activities.

### 4.1 Dataset

We utilized a dataset [22] containing the exchange data belonging to 5 different cryptocurrency exchange markets. To collect this dataset the CCXT library [21] has been used which provides the developers a unified way to get access to the data from a variety of cryptocurrency exchanges using the python programming language. Table 1 shows an instance of dataset.

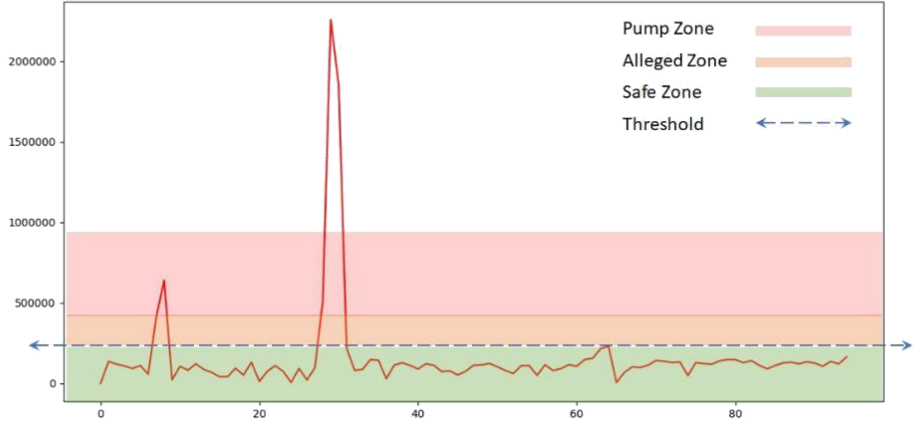

**Figure 7: Distance diagram for manual threshold setting with respect to initial frame.**

**Table 1: A sample instance of dataset.**

| Timestamp | Open | Low | High | Close | Trading Volume |
|---|---|---|---|---|---|
| 4/20/2018 1:00 | 0.00012931 | 0. 0.00012914 | 0.00012982 | 0.00012933 | 24057 |

**Table 2: Number of detected P & D activities on top 10 symbols ranked by number of alleged pump and dumps.**

| Exchange | Symbol pair | Alleged P&D | Distance based | Density based | Hybrid |
|---|---|---|---|---|---|
| Lbank | DBC/NEO | 13 | 8 | 5 | 9 |
| Kucoin | CAPP/BTC | 11 | 7 | 4 | 11 |
| Lbank | TKY/ETH | 10 | 10 | 3 | 11 |
| Bittrex | DCT/BTC | 10 | 2 | 4 | 4 |
| Bittrex | BRX/BTC | 9 | 5 | 7 | 8 |
| Binance | MDA/ETH | 9 | 3 | 4 | 5 |
| Bittrex | EMC/BTC | 8 | 10 | 4 | 11 |
| Kucoin | ADB/BTC | 7 | 2 | 3 | 3 |
| Bittrex | GNT/ETH | 7 | 4 | 7 | 7 |

## 4.2 Evaluation

To evaluate the proposed method, we use the ground truth provided in [4]. Our reason is to test capability of density-based, distance-base and hybrid methods to find anomalous instances. Table 2 shows top 10 symbols ranked by number of pump and dump. The number of P & D events is used as a ground truth to compare with number of detected P & D events using the distance-based, density-based and hybrid point anomaly methods. To the best of our knowledge, such an evaluation has not been reported in previous works.

4.2.1 *Test Case 1*. We test our proposed method on TKY/ETH exchange data belonging to Lbank exchange market. The ground truth shows 10 strange trading activities as being the result of a P & D. In this case, warning signals of corresponding price and volume spikes happen at different values as shown in figure 9. It's very difficult to find a reasonable threshold for price and volume. Projection of data into 2D using proposed method allows us to reduce the number of thresholds to only one. Part (a) of figure 8 shows the projected data into 2D. Setting the threshold based on this figure is much easier. However, processing the distance histogram with respect to the global centroid enables us to find the threshold automatically. Part (b) of figure 8 shows that the detected threshold is 4000 since the sparse region starts at this value. Counting the sparse region values reveals that there is 10 P & D activities as reported in Table 2. Figure 12 shows that, the data is more sensitive to distance since it is stretched in X direction. Table 2 confirms this expectation showing that distance-based approach has detected 10 anomalous instances and density-based approach has detected only 3 anomalous points which two of them have been common as tabulated in Table 3.

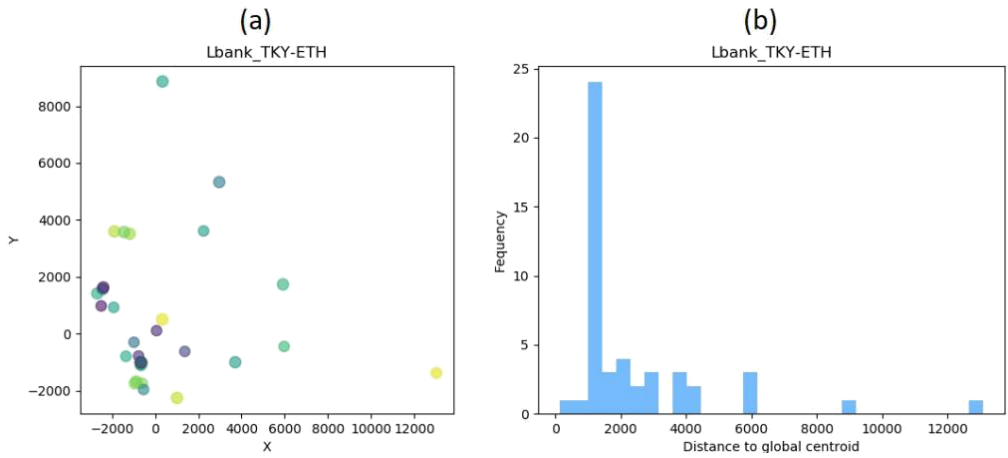

**Figure 8: Part (a): Projected TKY/ETH trading pair data into 2D. Part(b): corresponding distance histogram.**

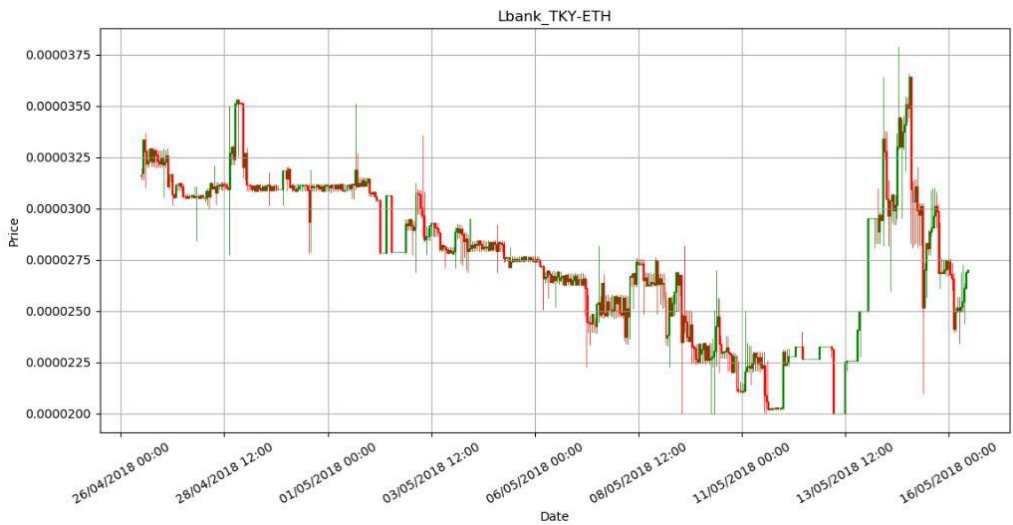

**Figure 9: Candlestick chart for the TKY/ETH trading pair.**

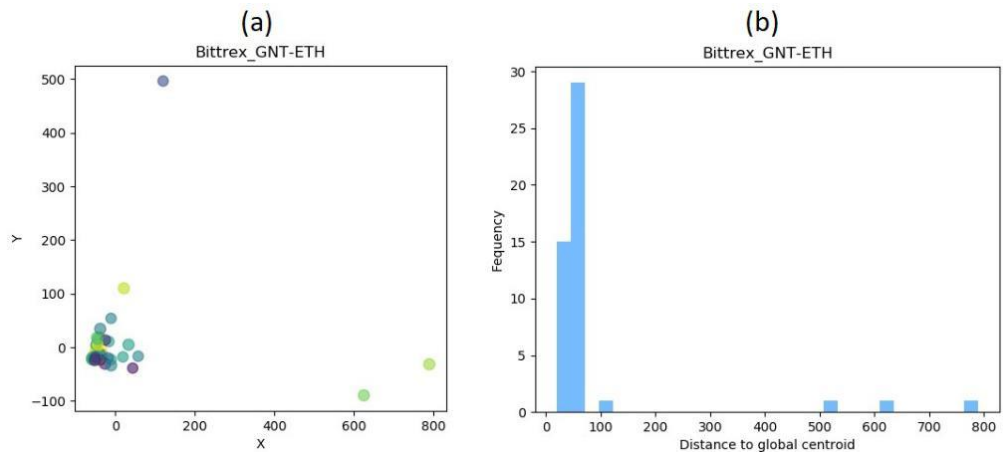

**Figure 10: Part (a): Projected GNT/ETH trading pair data into 2D. Part(b): Corresponding distance histogram.**

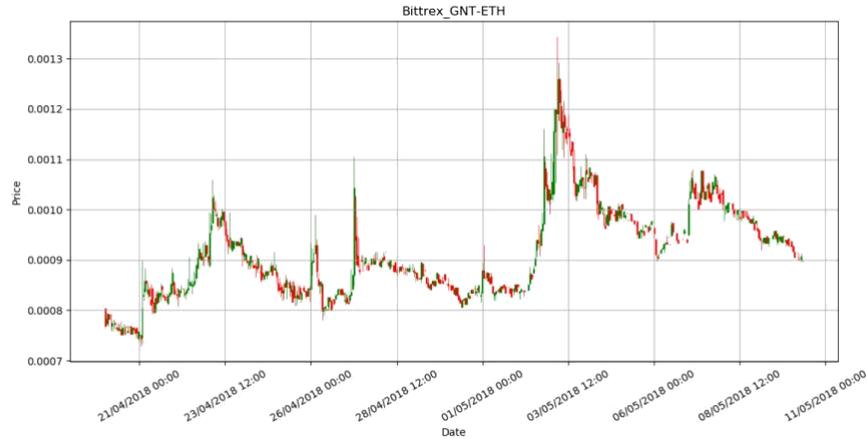

Figure 11: Candlestick chart for the GNT/ETH trading pair.

Table 3: Distance - Density statistics

| Exchange | Pair | Max Density | Detected Threshold | Common outliers | Impact |
|---|---|---|---|---|---|
| Lbank | DBC/NEO | 9 | 8000 | 4 | Distance |
| Kucoin | CAPP/BTC | 8 | 500 | 0 | Distance |
| Lbank | TKY/ETH | 8 | 4000 | 2 | Distance |
| Bittrex | DCT/BTC | 8 | 1400 | 2 | Density |
| Bittrex | BRX/BTC | 8 | 4 | 4 | Density |
| Binance | MDA/ETH | 8 | 210 | 2 | Density |
| Bittrex | RBY/BTC | 8 | 100 | 2 | Density |
| Bittrex | EMC/BTC | 8 | 18 | 3 | Distance |
| Kucoin | ADB/BTC | 8 | 3000 | 2 | Density |
| Bittrex | GNT/ETH | 7 | 95 | 4 | Density |

4.2.2 *Test Case 2*. We tested our proposed method on GNT/ETH exchange data belonging to Bittrex exchange market. The ground truth shows 7 strange trading activities as being the result of a P & D. In this case, warning signals of corresponding price and volume spikes happen at different values as shown in figure 11. It's very difficult to find a reasonable threshold for price and volume. Projection of data into 2D using proposed method allows us to reduce the number of thresholds to only one. Part (a) of figure 10 shows the projected data into 2D. Setting the threshold based on this figure is much easier. However, processing the histogram of points distances to the global centroid enables us to find the threshold automatically. Part (b) of figure 10 shows that the detected threshold is 95 since the sparse region starts at this value. Counting the sparse region values reveals that there is 4 P & D activities as reported in Table 2. Figure 13 shows that, the data is more sensitive to density since it is stretched in Y direction . Table 2 confirms this expectation showing that density-based approach has detected 7 anomalous instances and distance-based approach has detected only 4 anomalous points which all of them have been common between density-based and distance-based approaches as tabulated in Table 3.

Here is a summary of our experiments:

- Hybrid method shows the highest success rate comparing to distance-based and density-based approaches as shown in part (a) of figure 13.
- The overall success rate of hybrid method is higher than distance-based and density-based overall success rates by 23 percent and 30 percent, respectively.
- Density- Distance distribution diagram can reveal the impact of density and distance metrics on anomaly detection: If the distribution is stretched in X direction it's distance sensitive and if the distribution is stretched in Y direction it's density sensitive as shown in figure 12. Checking the Table 3 proves the stretch direction and impact relationship.
- The hybrid approach detects only 4 false positive instances. False positives are seen only in Lbank and Bittrex exchange markets as shown in part (f) of figure 13.
- 60 percent of top exchange pairs are density sensitive and 40 percent of them are distance sensitive as shown in part (d) of figure 13.
- Only 21 percent of detected outliers by density-based and distance-based approaches are common. It shows that, a combination of distance and density metrics as a hybrid approach can be more successful as shown in part (a) of figure 13.

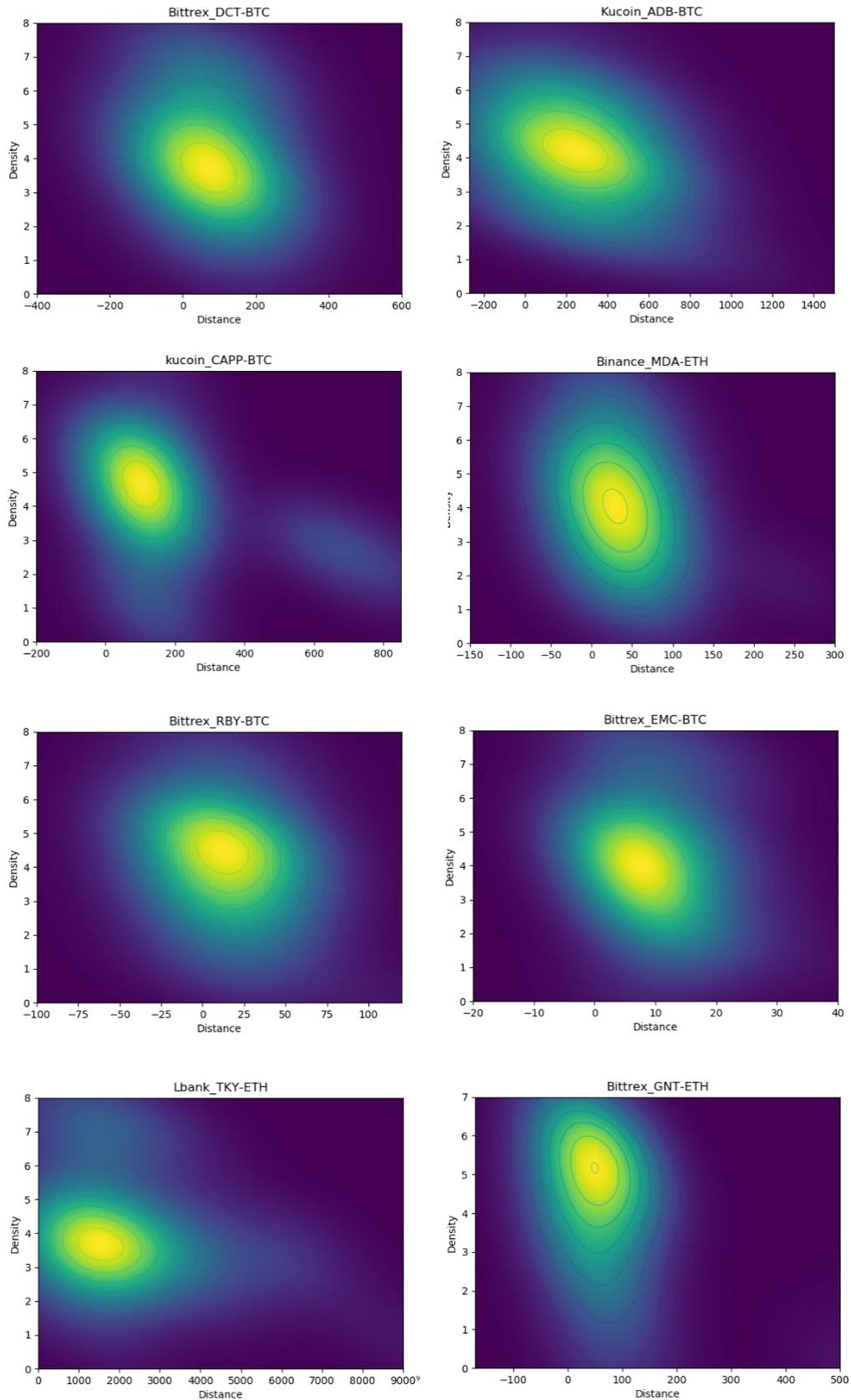

**Figure 12:** Density-Distance distribution diagram of top ranked exchange pairs.

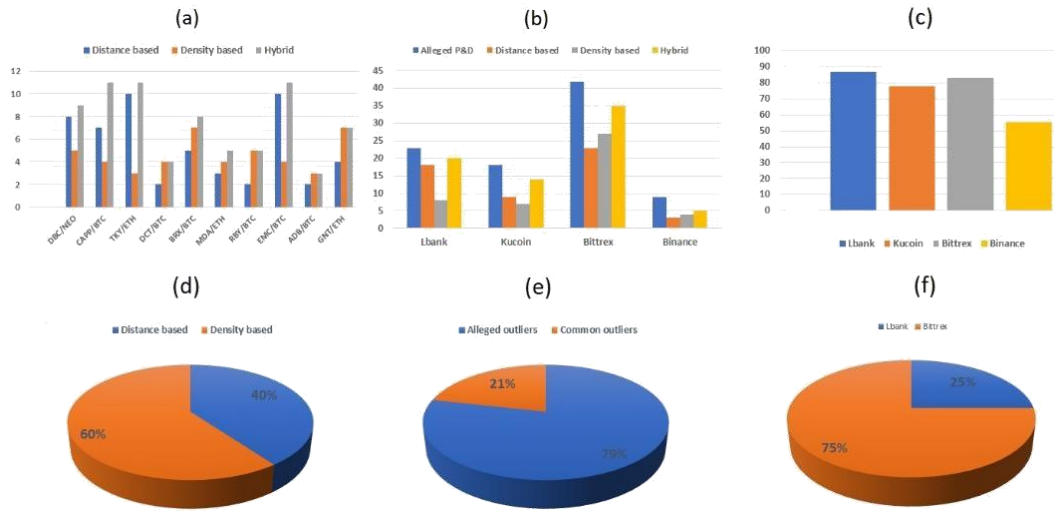

Figure 13: Experimental Results: Part(a),(b): Hybrid method versus distance-based and density-based methods. Part(c):Success rate in different exchange markets. Part(d): Density-Distance impact on top 10 exchange pairs. Part (e): the common outliers rate between distance-based and density-based methods. Part(f): The false positive rate.

- Highest hybrid success rate is seen in Lbank exchange mar-ket as shown in parts (b) and (c) of figure 13.

## 5 CONCLUSION

In this paper, we proposed a novel hybrid anomaly detection method for pump and dump detection. Contextual anomaly detection is challenging since it needs setting multiple thresholds. To Address this problem, we transformed the contextual anomaly problem into point anomaly problem via a novel method. First, we divided time series data into a set of different frames. Second, we concatenated all the data inside each frame to form a single high dimensional data per frame. Afterwards, we used distance and density metrics to find the point anomalies. We proposed a novel histogram processing approach for automatic distance threshold setting. We also proposed a novel approach to calculate the density score of each instance to detect the anomalies. Our experiments showed that, none of density based and distance-based approaches can obtain the best results. To solve this problem, we tested a hybrid approach by taking advantage of both distance and density metrics. The experimental results proved the superiority of hybrid P & D detection against distance-based and density-based approaches.